\documentclass[10pt,twocolumn,letterpaper]{article}

\usepackage[pagenumbers]{cvpr} 

\usepackage{graphicx}
\usepackage{amsmath}
\usepackage{amssymb}
\usepackage{booktabs}
\usepackage{makecell}

\usepackage[pagebackref,breaklinks,colorlinks]{hyperref}

\usepackage[capitalize]{cleveref}
\crefname{section}{Sec.}{Secs.}
\Crefname{section}{Section}{Sections}
\Crefname{table}{Table}{Tables}
\crefname{table}{Table }{Tabs.}

\def\confName{CVPR}
\def\confYear{2022}

\begin{document}

\title{
Multi-Forgery Detection Challenge 2022: \\
Push the Frontier of Unconstrained and Diverse Forgery Detection}

\author{Jianshu Li, Man Luo, Jian Liu, Tao Chen, Chengjie Wang, Ziwei Liu, Shuo Liu \\
Kewei Yang, Xuning Shao, Kang Chen \\
Boyuan Liu, Mingyu Guo \\
Ying Guo, Yingying Ao, Pengfei Gao \\
}

\maketitle

\begin{abstract}
   In this paper we present the Multi-Forgery Detection Challenge held concurrently with the IEEE Computer Society Workshop on Biometrics at CVPR 2022. Our Multi-Forgery Detection Challenge aims to detect automatic image manipulations including but not limited to image editing, image synthesis, image generation, image photoshop,~\emph{etc}. Our challenge has attracted $674$ teams from all over the world, with about $2000$ valid result submission counts. We invited the Top 10 teams to present their solutions to the challenge, from which three teams are awarded with prizes in the grand finale. In this paper, we present the solutions from the Top 3 teams, in order to boost the research work in the field of image forgery detection. 
\end{abstract}

\section{Introduction}
\label{sec:intro}
Image forgery refers to the set of automatic image manipulation techniques, for example, image editing, image synthesis, image generation, image photoshop, etc. In recent years, the abuse of image forgery techniques has raised enormous public concerns. In real-world digital identity authentication scenarios, image forgery techniques have been used by hackers to attack digital identity accounts. In other scenarios, hackers can also make any politicians talk whatever they want, or make any victims appear in any types of fake videos. Therefore, the detection of image forgery is an important technology for the security of AI systems.

\begin{figure}
  \centering
    \includegraphics[width=0.45\textwidth]{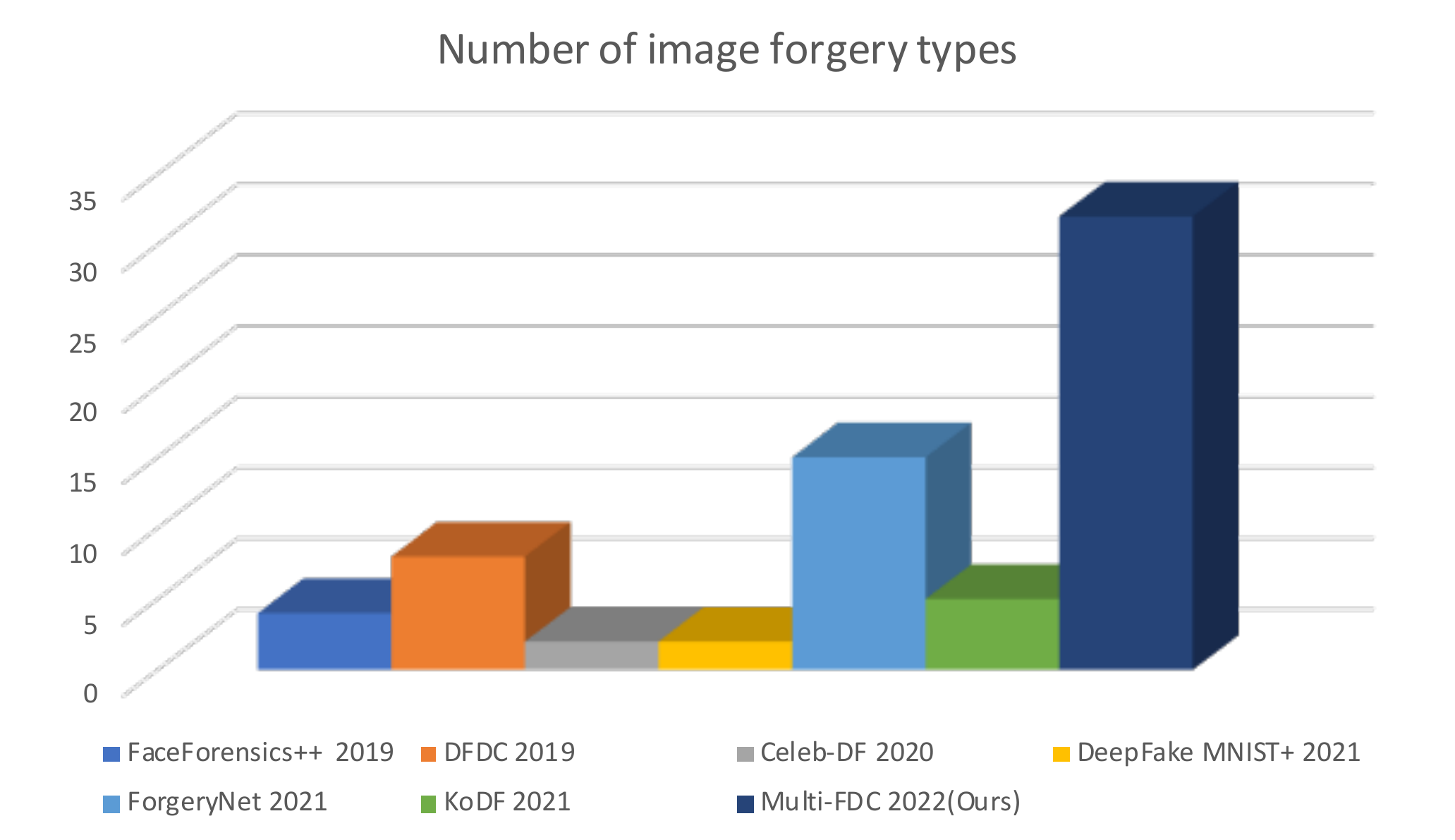}
    \caption{Number of image forgery types of different works in the literature.}
  \label{fig:num}
\end{figure}

 In the literature of image forgery detection, existing image forgery detection datasets are usually directly downloaded from YouTube or crafted by only a few popular image forgery software. Different from them, in our challenge we propose a new dataset named Multi-Forgery Detection Challenge (Multi-FDC) dataset. It covers more real-world scenes and a wide range of image forgery techniques such as face swapping, reenactment, attribute editing, fully synthesis, artificial PS, etc. The numbers of types of image forgery, compared with other works in the literature, are shown in Figure~\ref{fig:num}. We can see from the figure that in the current literature, the number of image forgery types is usually limited, and most works focus on a few mainstream image forgery types. However, most of the undetected forgeries are usually from unseen types. Thus it is important to consider multiple types of forgery when designing detection models. In our challenge, we provide a dataset with the most number of forgery types, in order to address the unconstrained and diverse forgery detection problem. Here we are not trying to enumerate all types of image forgery techniques in the literature, but we are advocating that modeling the unseen types of image forgery is essential for the task of image forgery detection. This challenge aims to encourage researchers around the world to develop innovative and generalized technologies that can defend more kinds of real-world image forgery attacks at the same time.

The rest of the paper is organized as follows. We will first demonstrate the setup of our challenge, and then present the details of the solutions from Top 3 teams. After that we will discuss the results from the teams and conclude the challenge. 

\section{Challenge Setup}
\subsection{Organizers}
Our challenge was hosted in conjunction with CVPR 2022 and 17-th IEEE Computer Society Workshop on Biometrics 2022. The organizers are Ant Group, Nanyang Technology University and the China Academy of Information and Communications Technology. The technical program was hosted on Alibaba Cloud TIANCHI platform.

\subsection{Datasets}
 In our challenge, we proposed a new dataset named Multi-Forgery Detection Challenge (Multi-FDC) dataset. The total size of the Multi-FDC dataset is about $1$ million. The real images in  Multi-FDC are mainly from publicly available datasets and images from online resources. For the forged images, we utilize a variety of forgery tools, mainly consisting of the following: face swapping, face reenactment, facial attributes editing, face synthesis, artificial PS, and other miscellaneous methods. The dataset in our challenge include over $30$ types of image forgeries, so that the participants have ample design space to model forgery types. The  number of images in the proposed Multi-FDC dataset is shown in Table~\ref{tab:num}. 

\begin{table}[h]
  \centering
  \begin{tabular}{@{}lcc@{}}
    \toprule
    Multi-FDC & Real images & Forged images \\
    \midrule
    training set & 99,878 & 499,831\\
    validation set & 19,986 & 20,000\\
    public testing set   & 104,641 & 98,993\\
    hidden testing set   & 31,818 & 23,721\\
    \bottomrule
  \end{tabular}
  \caption{The breakdown of number of images in the Multi-FDC dataset. }
  \label{tab:num}
\end{table}


Since the forgery of faces may cause more threats to AI systems, in our challenge we focus more on the face region rather than the background areas. As a result, all images in our dataset are aligned and cropped to $512 \times 512$, where the ratio of face regions is about $0.6\sim 0.7$. Moreover, we also concern about the generalization performance of the algorithm. Thus the testing sets contains new and unseen forgery types compared to the training and validation sets, in order to measure the generalization capability of  forgery detection models.

\subsection{Processes}
The task in our challenge is to judge whether an input image is forged or not. The whole challenge was divided into three phases,~\emph{i.e.} Phase 1, Phase 2, and Phase 3.

In Phase 1, only the training and validation sets are released. The forgery detection model can only be trained on the training set, with ImageNet pre-trained model weights. Data from external sources are not allowed in the training process. However, some image processing methods, such as face detection and alignment, face enhancement from the training dataset, are allowed to be used in the challenge. The validation set is also released, which can be used by the participants to improve the model performance and select the best model. Phase 1 lasted for about two months to provide enough time to perform model training and validation. 

In Phase 2,  the public testing set was released. Participants can directly submit the predicted score of the testing set to the platform and get immediate feedback on the evaluation scores. Phase 2 only lasted for two days to avoid over-fitting of the testing set. 

Top 10 teams from the leader board during Phase 2 can advance to the final Phase 3. In Phase 3, codes and models are submitted together with technical reports, which are used to produce prediction scores on our hidden testing set. The final ranking will be based on the weighted score of the public testing set, the hidden testing set and the technical report, and the weights are $0.4$, $0.4$ and $0.2$, respectively.

\subsection{Evaluation Methods}
For the performance evaluation, we mainly use the Area under the Curve (AUC). AUC is defined as the area under the Receiver Operating Characteristic (ROC) curve, and the value range is generally between $0.5$ and $1$. To be specific, in our setting, True Positive (TP), True Negative (TN), False Positive (FP) and False Negative (FN) are defined as follows:
\begin{enumerate}
    \item TP: The forged images are recognized as  forged images
\item TN: The real images are recognized as  real images
\item FP: The real images are recognized as  forged images
\item FN: The forged images are recognized as  real images
\end{enumerate}

With that, the True Positive Rate (TPR) and False Positive Rate (FPR) are defined in Equation~\eqref{eqn:tpr} and Equation~\eqref{eqn:fpr}, respectively. 

\begin{equation}\label{eqn:tpr}
    \mathbf{TPR} = \mathbf{TP}/(\mathbf{TP}+\mathbf{FN})
\end{equation}

\begin{equation}\label{eqn:fpr}
    \mathbf{FPR} = \mathbf{FP}/(\mathbf{FP}+\mathbf{TN})
\end{equation}

The ROC curve is essentially the TPR~\emph{v.s.} FPR curve, and AUC is the area under this curve.  To further assess and analyze the models, TPR at lower FPR, such as $1e-2$, $5e-3$, $1e-3$ will also be used as auxiliary metrics. However, the rankings will still be based on AUC.

\section{Solutions and Results}
Our challenge has attracted $674$ teams with about $2000$ valid submission counts. Our final validation set leader board has $106$ teams and the final test set leader board has $68$ teams. The Top $10$ teams are invited to the final phase, Phase 3, and $6$ teams actually participated in the final phase. The performance of the $6$ teams is shown in Table~\ref{tab:all}. 

\begin{table*}
  \centering
  \begin{tabular}{@{}lccc@{}}
    \toprule
    Rank & Team name & AUC on public test set & AUC on hidden test set \\
    \midrule
    1 & XiaoShiJiDao & 0.99386 &	0.98928 \\
    2 &	iDance	& 0.99132	& 0.97939	\\
    3 &	TrustCV	& 0.98793  &	0.97264	\\
    4 &	XXXXX  &	0.98098	& 0.92322	\\
    5 &	TingWoShuoXieXieNi &	0.97858	&0.91229	\\
    6 &	BuYiYangDeMeiNanZi &	0.96308&	0.94633	\\
    \bottomrule
  \end{tabular}
  \caption{The overall performance of the top ranking teams on the public test set and the hidden test set.}
  \label{tab:all}
\end{table*}

In the following subsections, we will present the solutions of the Top $3$ teams in our challenge. 

\subsection{Solution of the First Place}
\begin{itemize}
    \item Solution title: \textit{DAME: Data Augmentation and Model Ensemble for Generalized Face Forgery Detection}
    \item Team Name: \textit{XiaoShiJiDao}
    \item Team members: \textit{Kewei Yang, Xuning Shao, Kang Chen}
\end{itemize}

\subsubsection{General Method Description.}
The champion team proposed a robust and generalized method for forgery detection. It has two main components: \textbf{(1) data extension: } to improve the generalization, a data augmentation strategy is developed that extends the training data with six types of synthesized forgeries. \textbf{(2) model pool: } in which EfficientNet-B6 \cite{tan2019efficientnet} and MPViT-B \cite{lee2021mpvit} models are collected that perform better on the validation set.

\begin{figure}[h]
  \centering
  \includegraphics[width=0.85\linewidth]{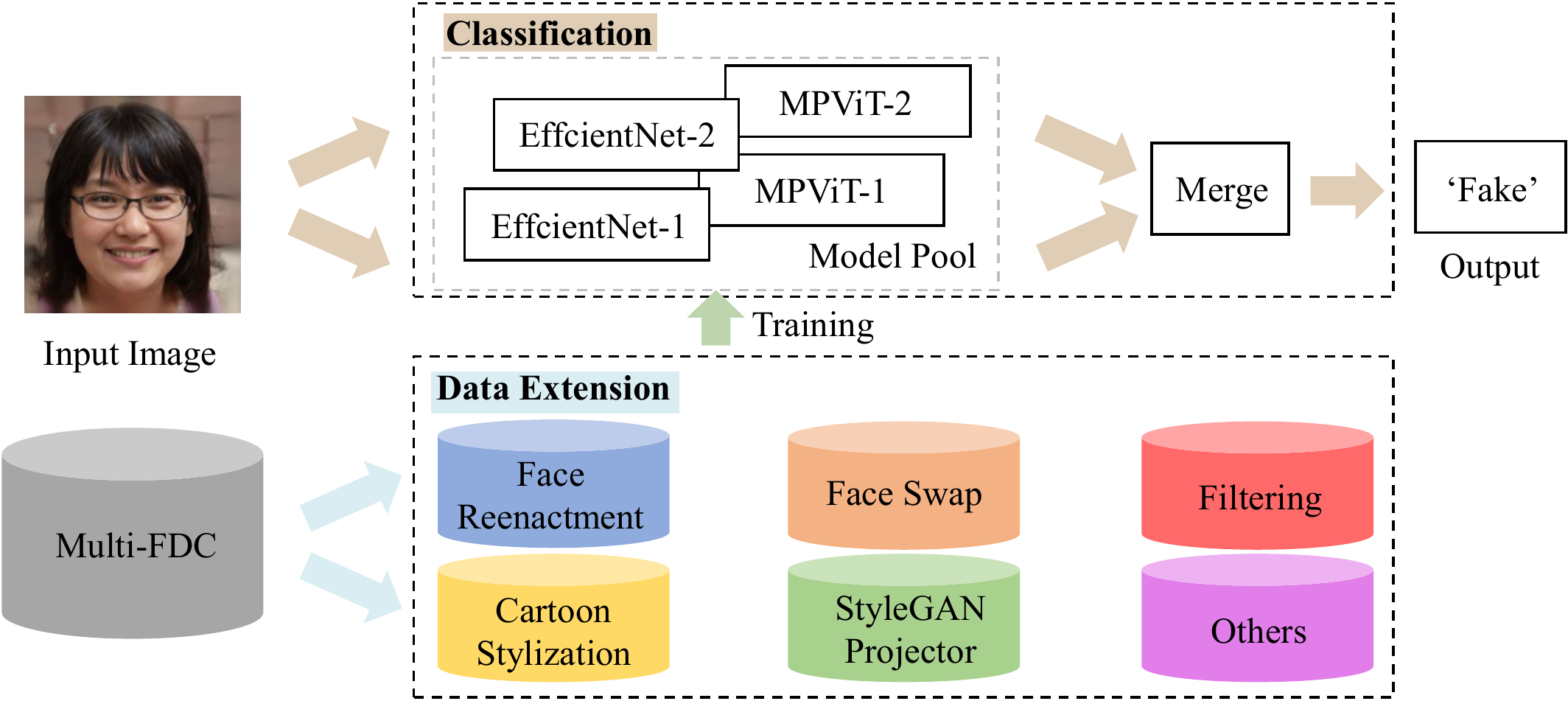}
  \caption{The DAME face forgery detection pipeline.}
  \label{Figure workflow}
\end{figure}

\begin{figure}[h]
  \centering
  \includegraphics[width=0.5\linewidth]{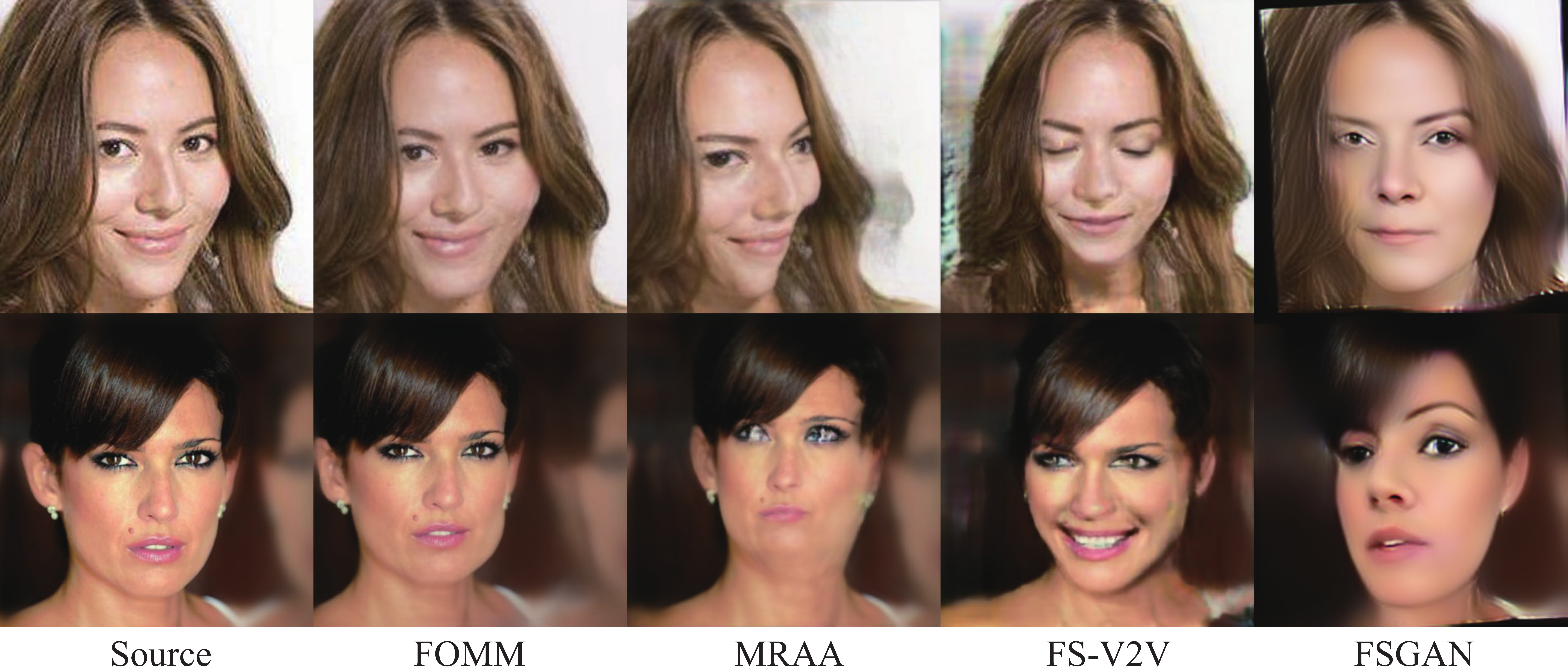}
  \caption{Sample images generated by face reenactment.}
  \label{Figure face_reenactment}
\end{figure}

\begin{figure}[h]
  \centering
  \includegraphics[width=0.6\linewidth]{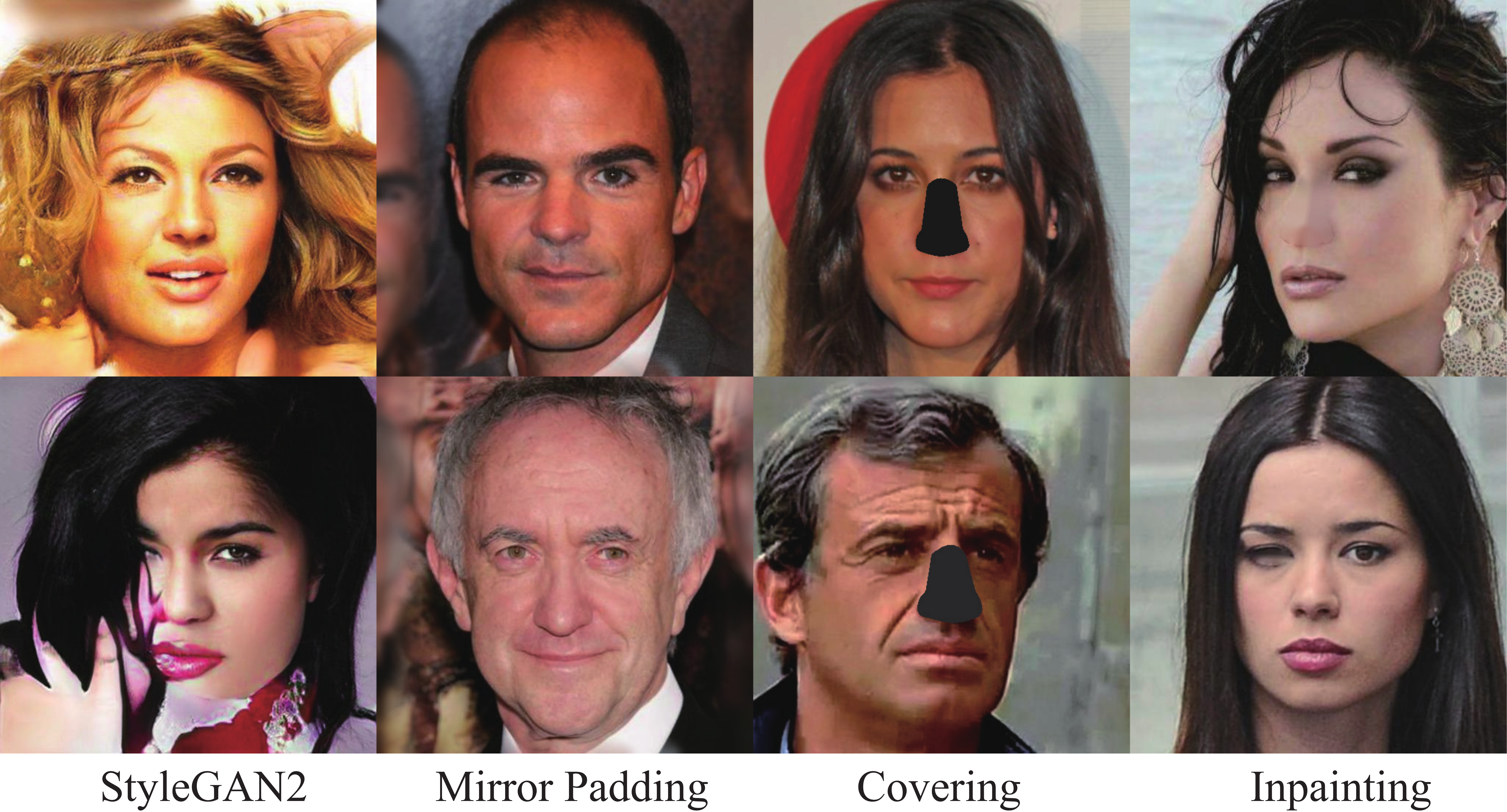}
  \caption{Sample images of other data extension methods.}
  \label{Figure others}
\end{figure}

\paragraph{Data extension:}
To address the domain generalization (DG) problem, the proposed solution managed to enrich the ``diversity" of forgeries by synthesizing augmented forgeries with six types of methods, including face reenactment (i.e., FOMM \cite{faceedit}, MRAA \cite{siarohin2021motion}, FS-V2V \cite{wang2019few} and FSGAN \cite{nirkin2019fsgan}), face swap (i.e., FSGAN \cite{nirkin2019fsgan}, face morphing \cite{steyvers1999morphing} and 3DMM entire/partial \cite{blanz1999morphable}), image filtering (i.e., Kuwahara filter \cite{kyprianidis2009image}, Bilateral filter \cite{banterle2012low}, xDoG filter  \cite{winnemoller2012xdog}, and stippling filter \cite{kopf2006recursive}), cartoon stylization (i.e., SPatchGAN \cite{shao2021spatchgan} and U-GAT-IT \cite{kim2019u}), StyleGAN projector (i.e., PSP \cite{tov2021designing} and E4E \cite{richardson2021encoding}), and others (i.e., StyleGAN2 \cite{styleganv2}, image covering and image inpainting \cite{bertalmio2001navier}). These methods cover a wide range of image manipulation techniques using traditional filtering approaches, computer graphics-based approaches, and learning-based approaches. Some sample images are shown in Figure ~\ref{Figure face_reenactment}~$\sim$ Figure ~\ref{Figure others}.

\begin{figure}[h]
  \centering
  \includegraphics[width=0.6\linewidth]{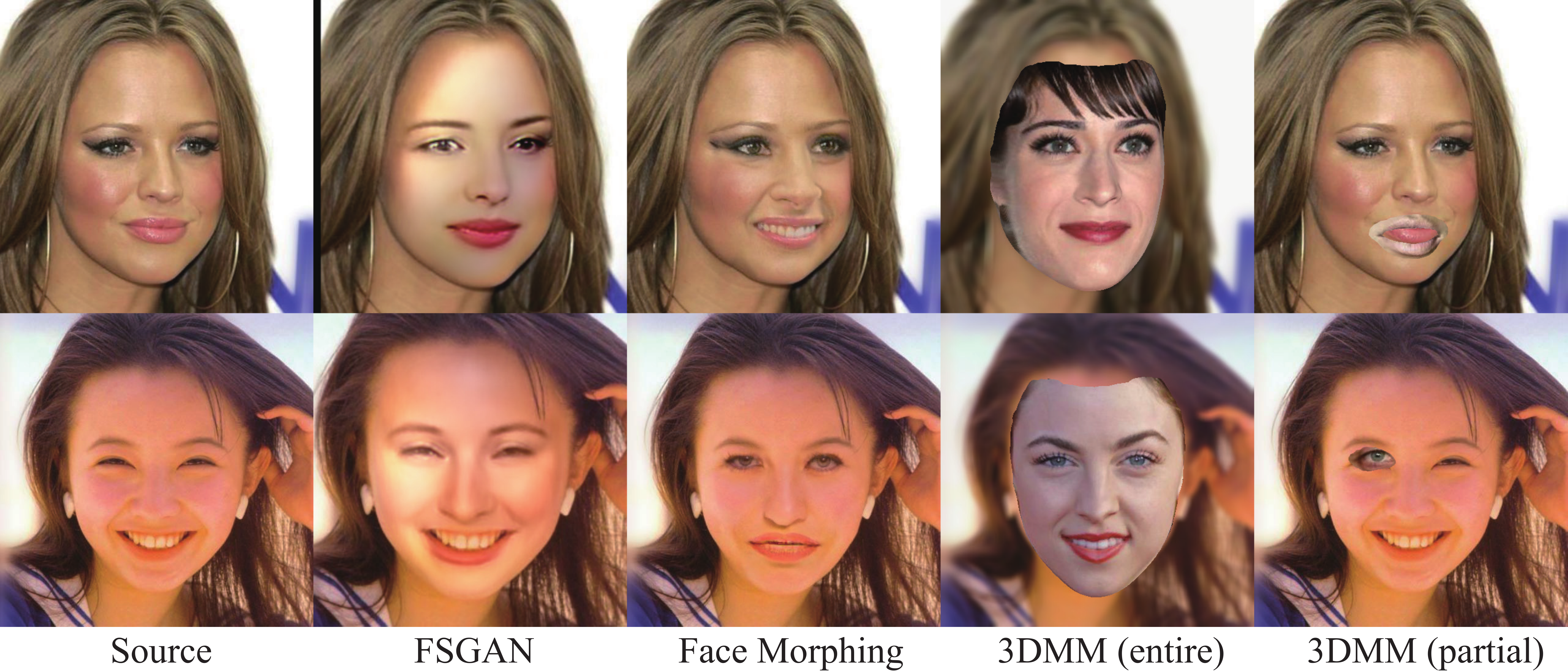}
  \caption{Sample images generated by face swap. ``Entire" means replacing the entire face area of the source image, and ``partial" means only replacing a random facial part. Both the source and target images are from the training set of Multi-FDC.}
  \label{Figure face_swap}
\end{figure}

\begin{figure}[h]
  \centering
  \includegraphics[width=0.6\linewidth]{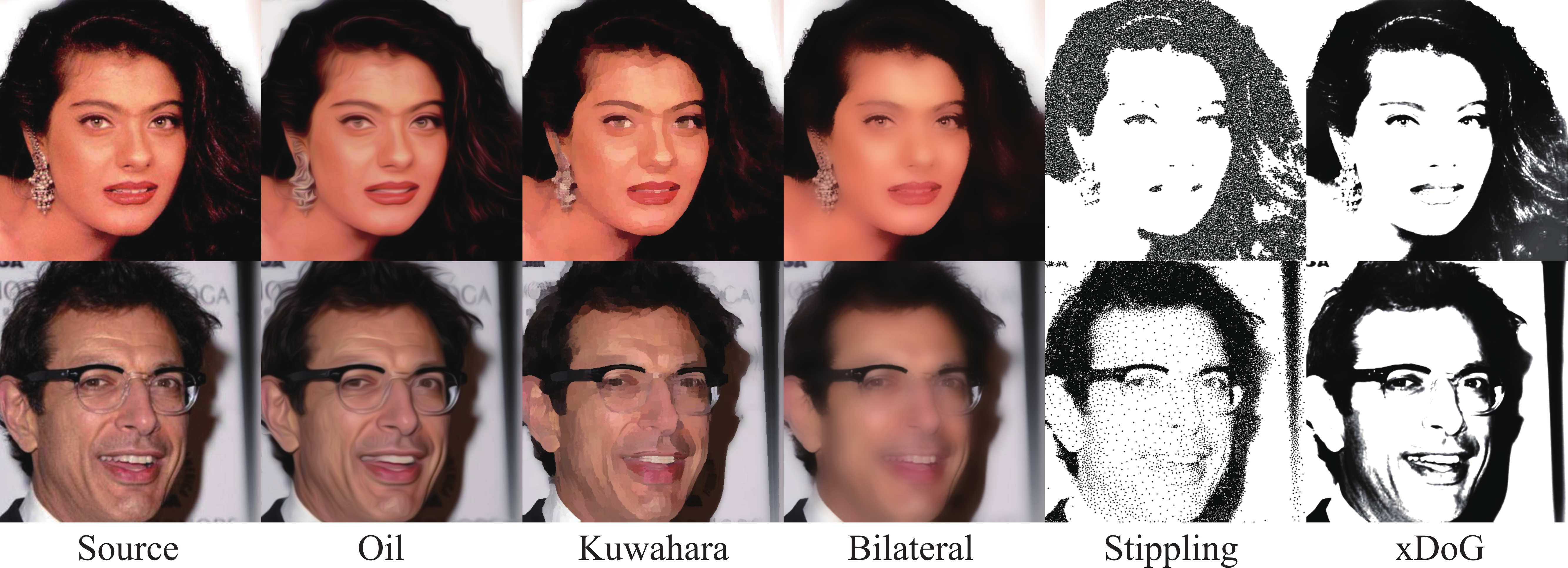}
  \caption{Sample images generated by image filtering.}
  \label{Figure image_filtering}
\end{figure}

\begin{figure}[h]
  \centering
  \includegraphics[width=0.6\linewidth]{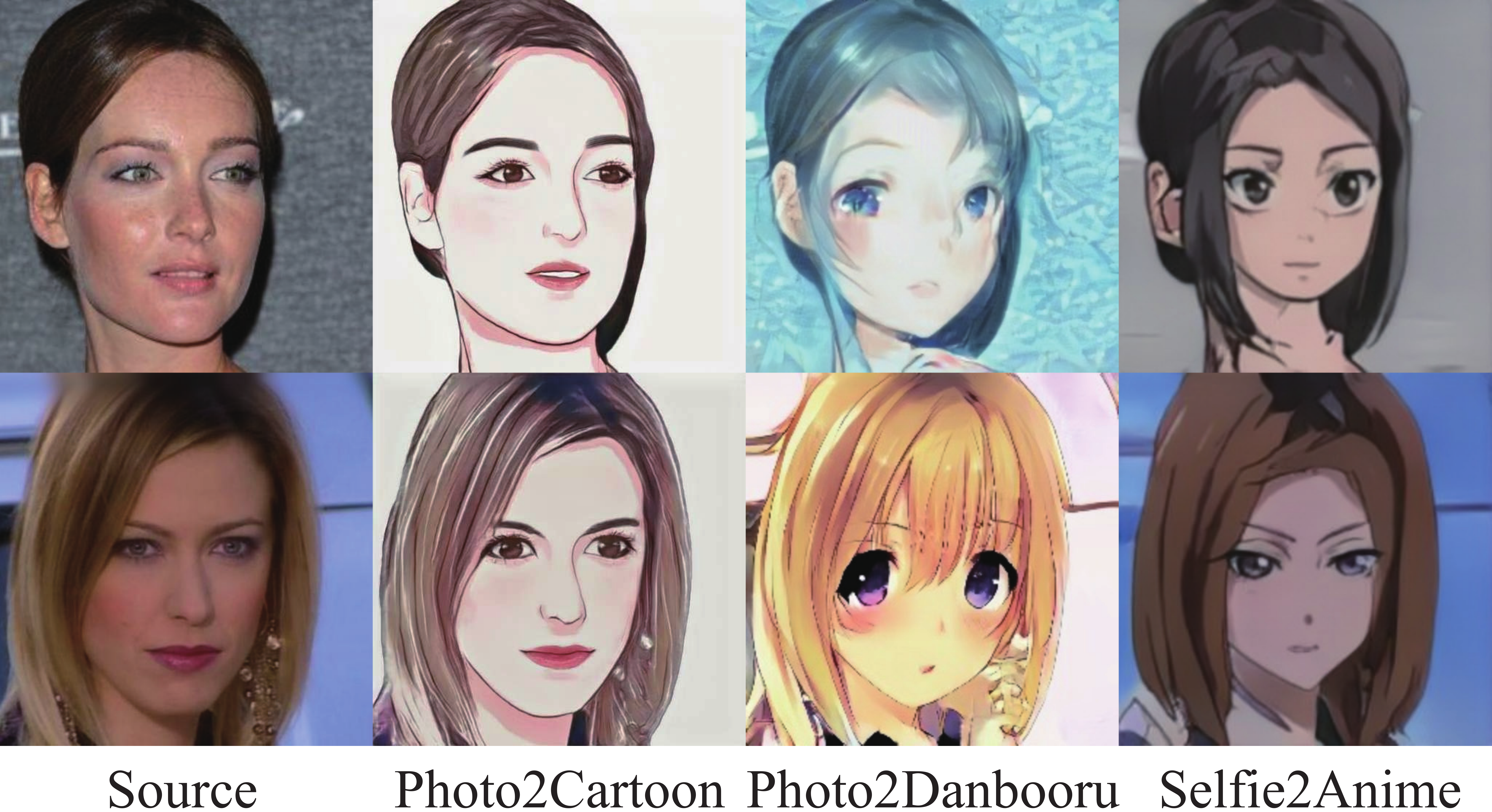}
  \caption{Sample images generated by cartoon stylization.}
  \label{Figure cartoon_stylization}
\end{figure}

\begin{figure}[h]
  \centering
  \includegraphics[width=0.6\linewidth]{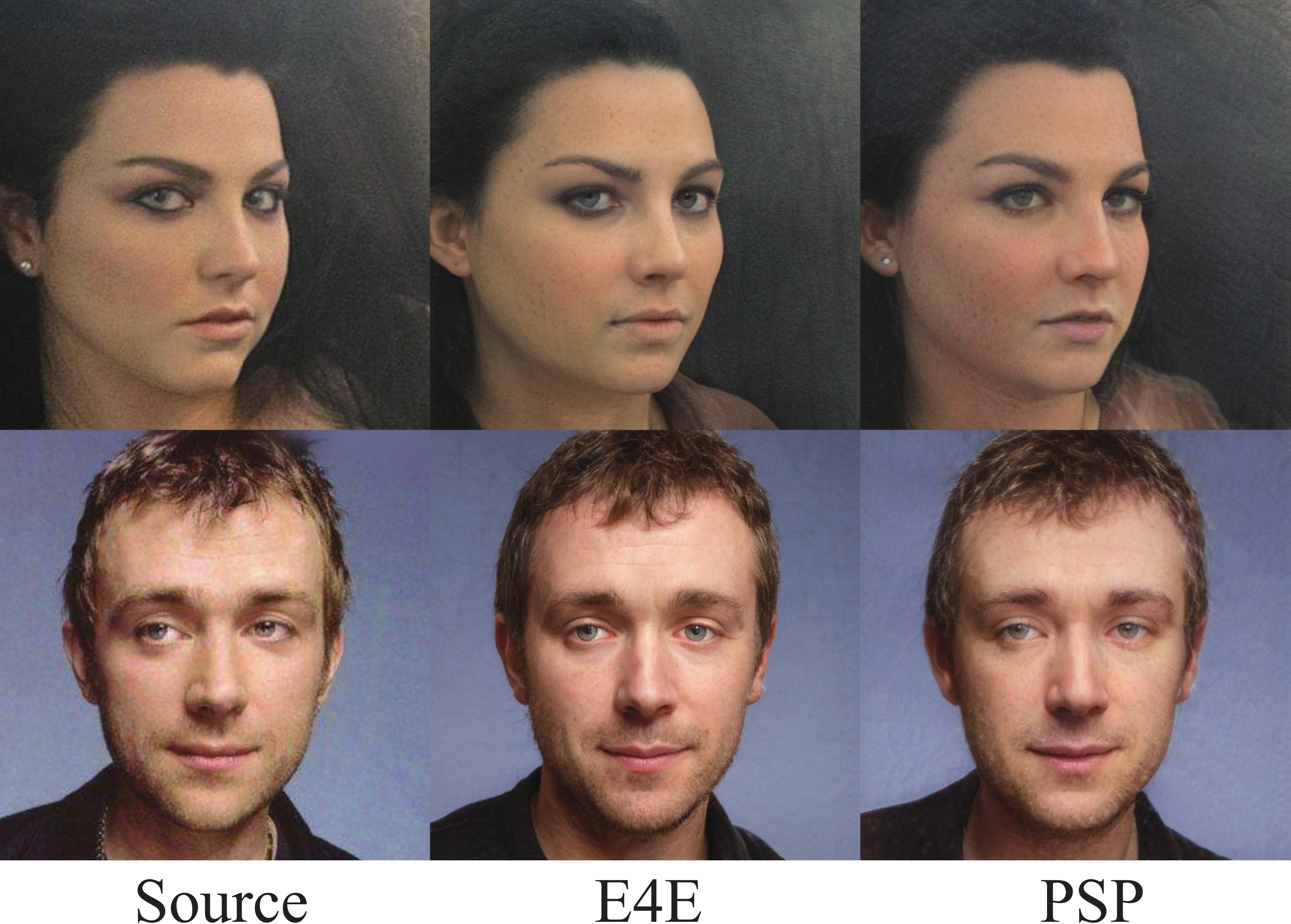}
  \caption{Sample images generated by StyleGAN projector.}
  \label{Figure styleGAN_encoder}
\end{figure}

\begin{figure*}[t]
  \centering
  \includegraphics[width=0.8\linewidth]{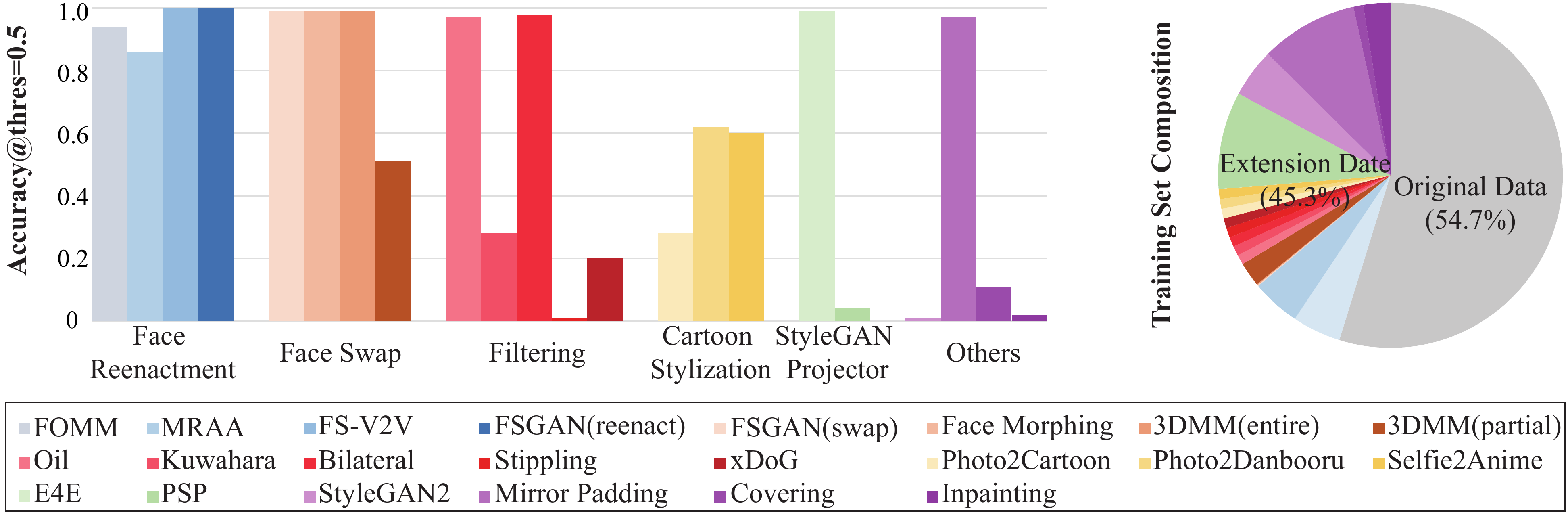}
  \caption{Left: accuracy of the baseline model on the different manipulation methods. Right: composition of the training set.}
  \label{Figure data_statistic}
\end{figure*}

The final training set consists of two parts: 600k original Multi-FDC images and 400k images from their extension data, as shown in Figure ~\ref{Figure data_statistic}. To select extended images with better potential to improve the detection performance, a baseline model is trained on the Multi-FDC dataset using EfficientNet-B6 \cite{tan2019efficientnet} and the error-prone images (e.g., roughly any fake image with a forgery score $<= 0.98$) are added to the training set. 

\paragraph{Model pool:} 
More than 50 models are trained based on different training set configurations, iterations and backbones and it is found that the models based on EfficientNet-B6 \cite{tan2019efficientnet} and MPViT-B \cite{lee2021mpvit} outperform the other backbones on the validation set. The final model pool consists of 4 models (e.g., 3 EfficientNet-B6 \cite{tan2019efficientnet} and 1 MPViT-B \cite{lee2021mpvit}) with different configurations, Table ~\ref{Table.model_pool} highlights some of them. During classification, these models are merged by averaging their respective predicted forgery scores, which can improve the accuracy and generalization performance.

\begin{table}[t]
\centering
\resizebox{\linewidth}{!}{%
\begin{tabular}{c | c  | c}
\hline
\textbf{Model} & \textbf{Training Set} & \textbf{Max Iteration}\\
\hline
\hline
\makecell[c]{EfficientNet-B6\\(Baseline)} & \makecell[l]{mfdc}  & 500k\\
\hline
\hline
\textbf{EfficientNet-B6-1}       & \makecell[l]{mfdc,fo,mr,fgs,c,mp,sg,e4e,psp,3e,3p,o,k,b,in} & 700k\\
\hline
\textbf{EfficientNet-B6-2}       & \makecell[l]{mfdc,fo,mr,fgs,c,mp,sg,st,e4e,psp,3e,3p,o,k,b} & 500k\\
\hline
\textbf{EfficientNet-B6-3}       & \makecell[l]{mfdc,fo,mr,fgs,c,mp,sg,st,e4e,psp} & 500k\\
\hline
EfficientNet-B6-4       & \makecell[l]{mfdc,fo,mr,fgs,c,mp,sg,st,e4e,psp,3e,3p,o,k,b,in} & 500k\\
\hline
\textbf{MPViT-B-1}       & \makecell[l]{mfdc,fo,mr,fgs,c,mp,sg,st,e4e,psp} & 500k\\
\hline
MPViT-B-2       & \makecell[l]{mfdc,fo,mr,fgs,c,mp,sg,st,e4e,psp,3e,3p,o,k,b,in,p2c,p2d,s2a} & 500k\\
\hline
XceptionNet \cite{chollet2017xception}   & \makecell[l]{mfdc,fo,mr,fgs,c,mp,sg,in} & 500k\\
\hline
ResNet-RS-420 \cite{bello2021revisiting}   & \makecell[l]{mfdc} & 500k\\
\hline
Swin-T-v2 \cite{liu2021swin}   & \makecell[l]{mfdc} & 500k\\
\hline
\end{tabular}
}
\caption{Configurations of different models. In the table, mfdc stands for the original Multi-FDC dataset, and fo, mr, fgs, c, mp, sg, st, e4e, psp, o, k, b, in, 3e, 3p, p2c, p2d, s2a stand for the extension data using FOMM, MRAA, FSGAN (swap), Covering, Mirror Padding, StyleGAN2, Stippling, E4E, PSP, Oil, Kuwahara, Bilateral, Inpainting, 3DMM (entire), 3DMM (partial), Photo2Cartoon, Photo2Danbooru and Selfie2Anime. The text in bold type indicates the selected models.}
\label{Table.model_pool}    
\end{table}

\subsubsection{Implementation details.}
\paragraph{Training description:}
During training,  the real images are repeated several times to overcome the data imbalance problem and all models are trained by minimizing the cross-entropy loss. An Adam \cite{kingma2014adam} optimizer with $\beta_1=0.9$ and $\beta_2=0.999$ is used. All models are initialized using the ImageNet-1K pre-trained models and trained with a batch size of 32. A linear learning rate scheduler is used for EfficientNet-B6 \cite{tan2019efficientnet}, and the initial learning rate is set to 0.0001. For MPViT-B \cite{lee2021mpvit}, CosineAnnealingWarmRestarts learning rate scheduler \cite{LoshchilovH17} is used. The initial learning rate is set to 0.0001, the number of iterations for the first restart (i.e., $T_0$) is set to 25k, and the increasing factor after a restart (i.e., $T_{mult}$) is set to 2.

In addition,  dropout \cite{srivastava2014dropout} and runtime augmentation strategies are adopted to reduce overfitting. The dropout ratio is set to 0.5, and the augmentations include randomly cropping, resizing, color jittering, and grayscale conversion.

\paragraph{Testing description:}
During testing, instead of only using the best model,  the forgery scores predicted by each model inside the model pool  is averaged, which can significantly improve accuracy over the best individual model.

\subsubsection{Generalization Analysis.}
Although recent studies in face forgery detection have yielded promising results when the training and testing face forgeries are from the same dataset, the problem remains challenging when one tries to generalize the detector to forgeries created by unseen methods \cite{chen2022self}. To address this domain generalization problem, the champion team focused on the data extension and improved the diversity of forgeries with six types of augmentation methods, which also proved to be very effective in this challenge. With the data extension paradigm, the corresponding fake data can be added to the training set for any new forgery methods. The domain generalization problem can be then turned into a continual learning problem \cite{DBLP:conf/coling/BiesialskaBC20} in practice.

\subsection{Solution of the Second Place}
\begin{itemize}
    \item Solution title: \textit{Multi-modal Multi-class Multi-forgery Detection}
    \item Team Name: \textit{iDance}
    \item Team members: \textit{Boyuan Liu, Mingyu Guo, Jiao Ran}
\end{itemize}
\subsubsection{General Method Description.}

In this challenge, an ensemble of 6 EfficientNet-b7 models are used under three training settings: \textbf{(1) baseline models}, \textbf{(2) multi-modal models} and \textbf{(3) multi-class models}.

\paragraph{Baseline models} are trained using raw images as input, without resizing or complex data augmentations. By this means, high-frequent noise patterns and GAN-fingerprints \cite{fridrich2012rich} are preserved. Only horizontal flipping and randomly erasing facial landmarks are  applied as data augmentations.

\begin{figure}[h]
    \centering
    \includegraphics[width=0.8\linewidth]{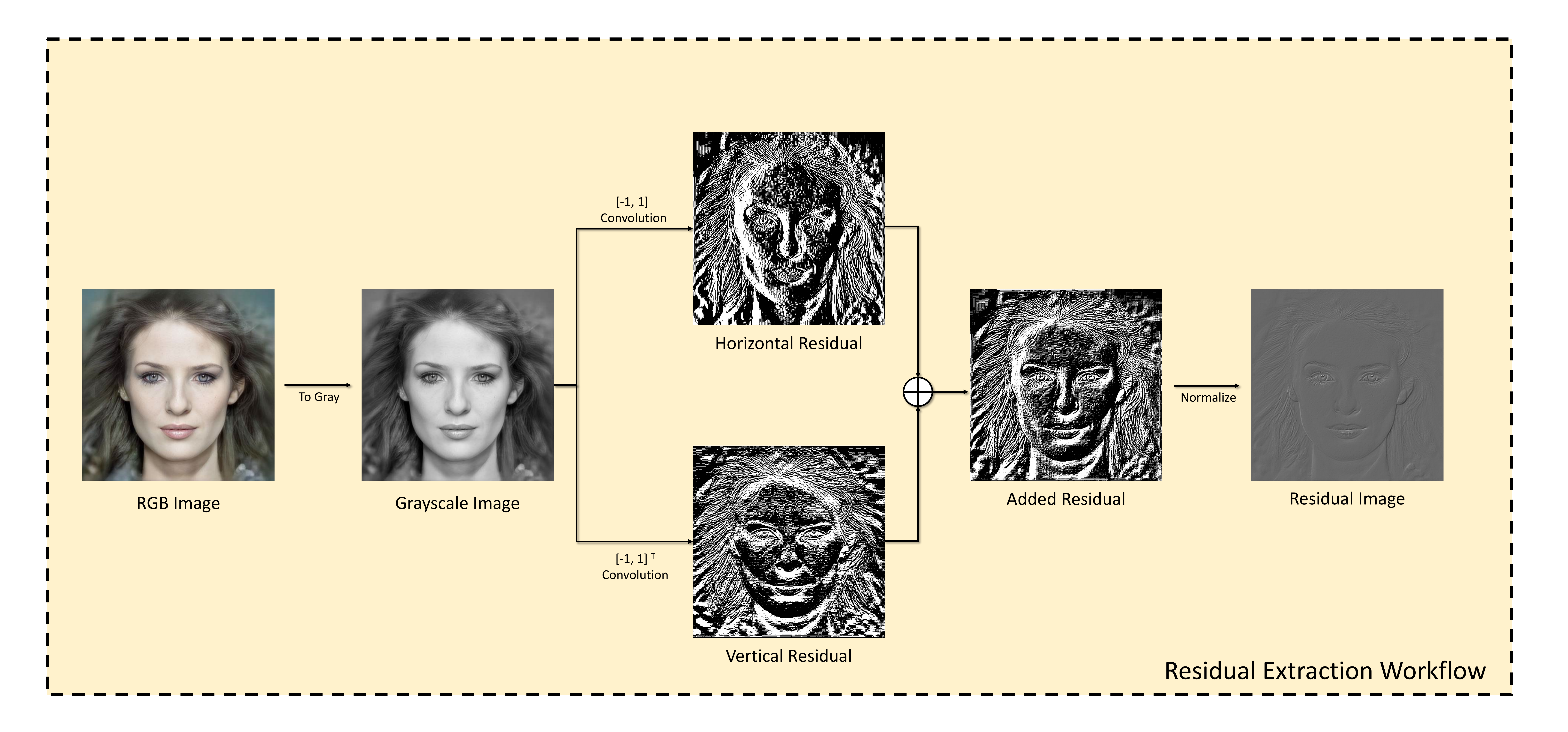}
    \caption{The workflow of image noise extraction}
    \label{fig:0}
\end{figure}

\paragraph{Multi-modal models} are trained using SRM (Spatial Rich Model) \cite{yu2019attributing} filtered images as inputs, using similar training setting as baseline models. Inspired by \cite{luo2021generalizing},   one simple filter-group followed by normalization as a novel modality is designed for multi-forgery tasks, which is applied on images before inputted into the backbone. To be specific, the  original image is converted into gray-scale image, and then two convolution operations are applied on the gray-scale image with kernel $[-1, 1]$ and $[-1, 1]^T$ respectively. The output image is normalized after pixel-wisely adding up the vertical residual image and the horizontal residual image. The workflow can be seen in Figure. \ref{fig:0}. 

\begin{figure}[h]
    \centering
    \includegraphics[width=0.6\linewidth]{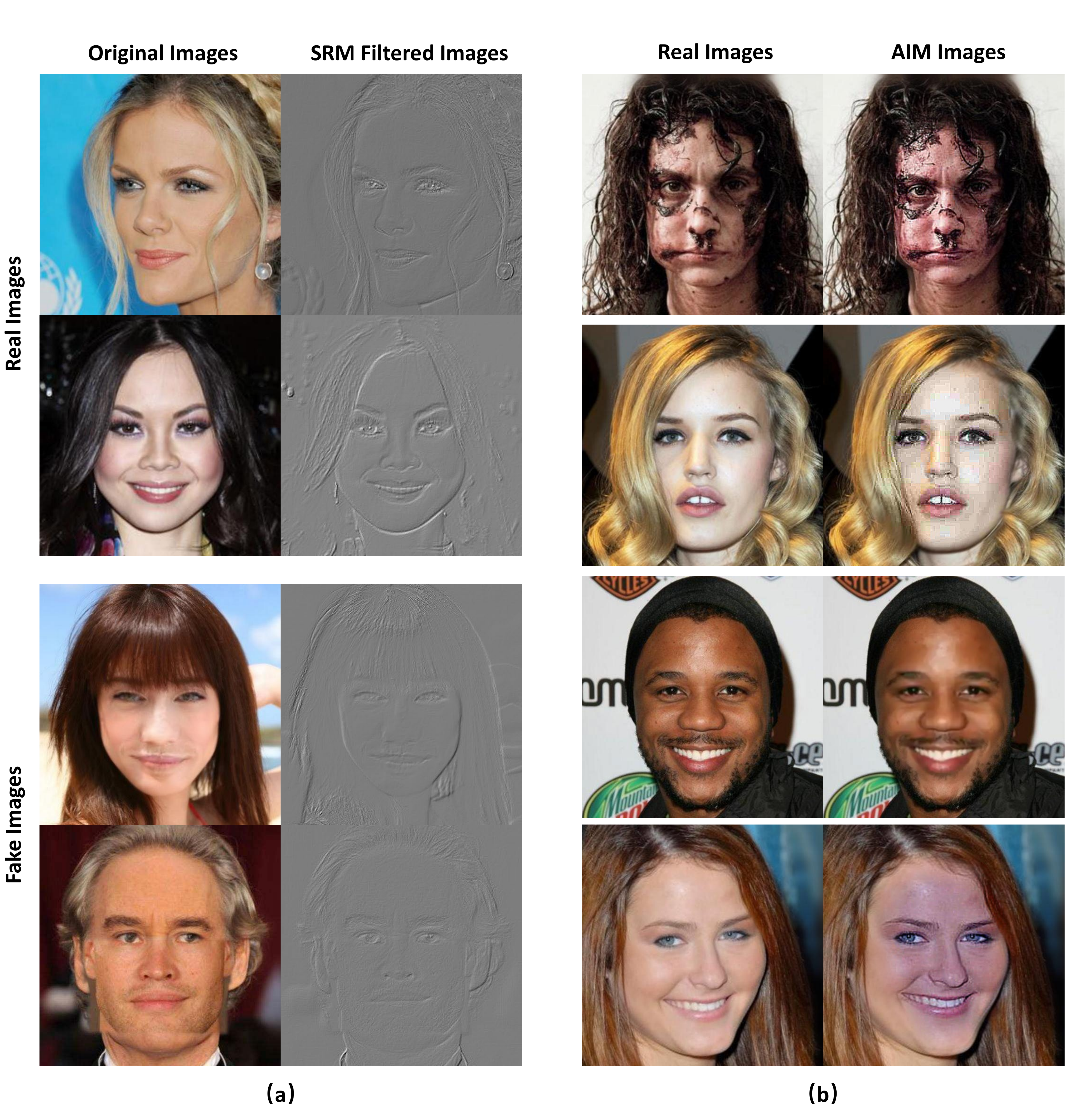}
    \caption{Examples of RGB images and SRM filtered images are shown in (a) and examples of real images and AIM images are shown in (b).}
    \label{fig:1}
\end{figure}

With such operations, the method can amplify image residuals containing noise patterns like CNN textures and GAN-fingerprints produced by generative networks. Besides, as color information is deprived by SRM filters, models can focus more on image residuals, ignoring irrelevant background color information. Representative SRM filtered images are shown in Figure. \ref{fig:1}.

\begin{figure}[h]
    \centering
    \includegraphics[width=0.75\linewidth]{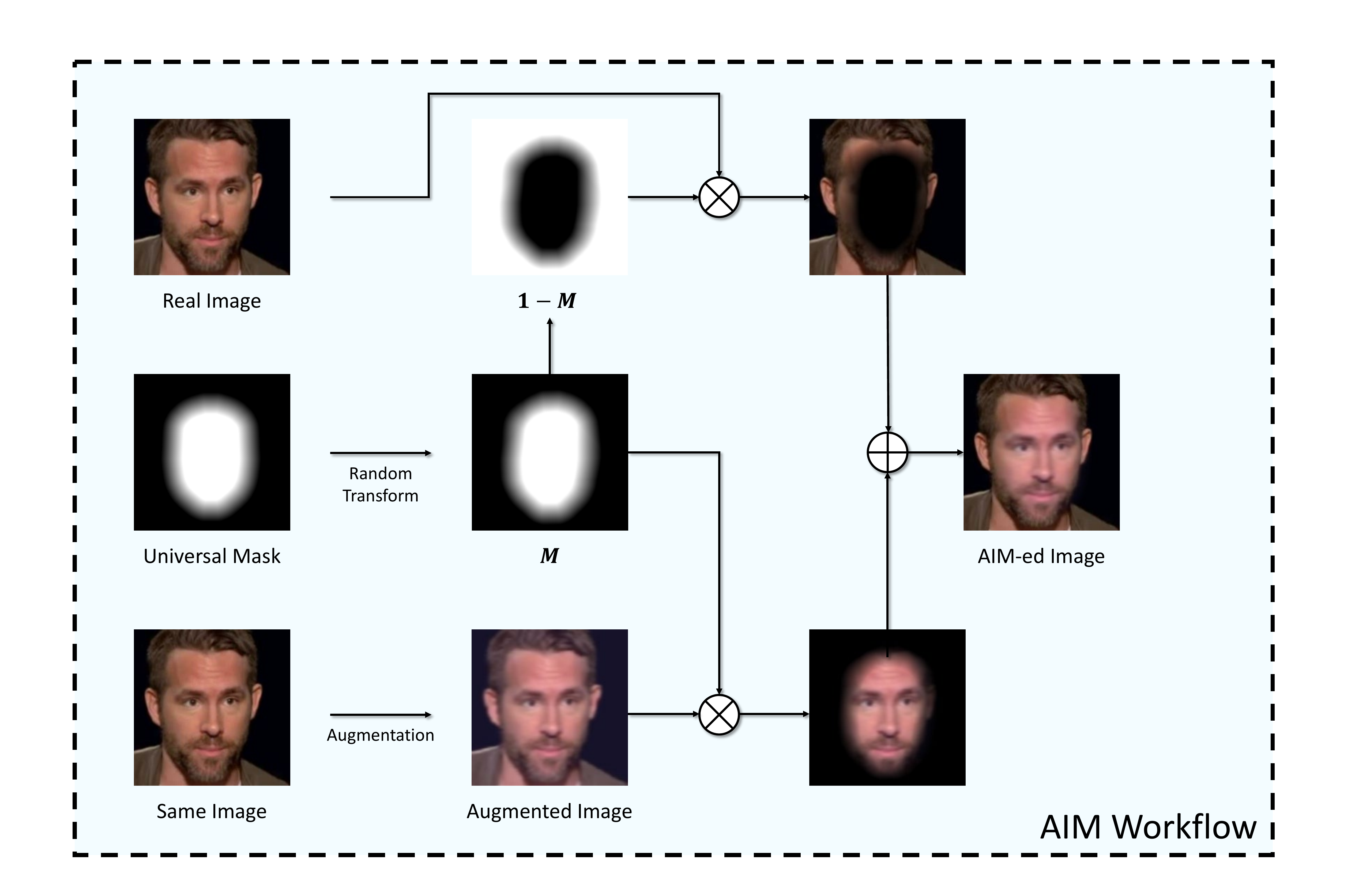}
    \caption{AIM Workflow.}
    \label{fig:2}
\end{figure}

\paragraph{Multi-class models} are trained under the same settings as baseline models, while assigned an extra class using the proposed fake sample generation module named AIM (Augmentation Inside Mask), that can on-the-fly generate fake samples during training. Given a real image, AIM apply complex data augmentation inside mask areas (face region), and output images with different noise patterns, colors, resolutions and landmark positions between face area and background. The workflow and AIM data examples are shown in Figure. \ref{fig:2}, Figure. \ref{fig:1}.(b), respectively. 

As a result of AIM, the noise pattern of fake images are different from real images. The authors asserted that, inevitably, most unseen forged images are generated by stitching target and source images. Therefore, the forged image contains different source features at different locations, whereas those of a pristine image must be consistent across all positions. Label 2 is assigned for those images with inconsistent noise pattern between face region and its background. In this way, as real images are labeled 0, seen forgeries are classified as label 1, and unseen forgeries are classified as label 2.

\subsubsection{Implementation details.}
\paragraph{AIM Workflow:}
To generate an AIM sample, a real image is first resized at a random scale before data augmentation is applied in the mask region, in order to produce noise pattern of different scale (e.g. JPEG blocking artifacts). To simulate different noise patterns between face and background, different noisy augmentation like image compression (JPEG compression and WEBP compression), down-sampling, Gaussian noises, Gaussian blur and sharpening are used with random scales. Besides, in some situations where the colors between the face region and the background are different,  RGB shift and color jitter are applied. Elastic transforms are also used to mimic facial landmark mismatch in some situations where image forgery are of low quality. It is worth noting that all the aforementioned augmentation are applied on the face region. To accomplish this, the augmented image is resized to the original size $512\times512$, and instead of calculating the convex hull of all facial landmarks,  a universal mask (shown in the middle of the left side of Figure \ref{fig:2}) is drawn to accelerate the generating procedure, as most images in the dataset are already aligned. To diversify the mask size and shape,  some random transformations are applied to the universal mask, including eroding, dilating and elastic transform. Finally the AIM image is calculated by combining the original image $I$ and the augmented image $I_a$
$$I_{AIM}= M\odot I_a + (1-M)\odot I $$

\begin{figure}[t]
    \centering
    \includegraphics[width=0.5\linewidth]{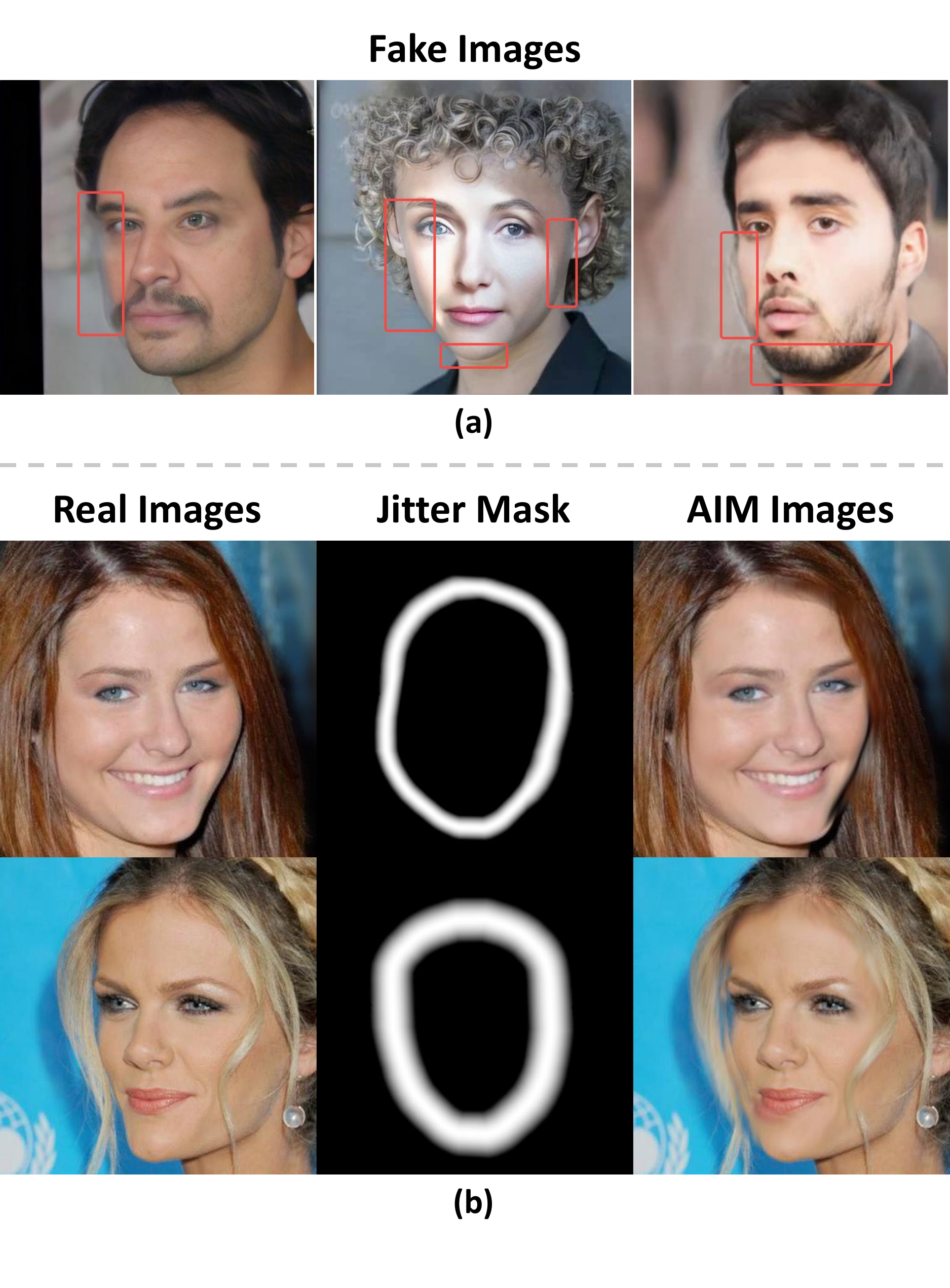}
    \caption{As shown in (a), the red boxes note that some fake images, despite forged in high fidelity, still expose some shadow-like artifacts on the edge of face region. To mimic such artifacts during training, the edge area of AIM mask is calculated and motion blur is applied inside mask area, to produce similar artifacts, shown in (b).}
    \label{fig:add}
\end{figure}

Apart from these,  in some cases the forged faces are eye-deceiving, but still some shadow-like artifacts are exposed around the edge area of face region, as shown in  Figure \ref{fig:add}(a). This may happen in some GAN-based face swapping techniques, probably in some inevitable circumstances where the face width between source faces and target faces varied. To cope with this, an improved version of AIM is proposed that produce shadow-like artifacts. As shown in Figure \ref{fig:add}(b), the jitter mask like Face X-ray \cite{facexray} is calculated as
$$M' = M \odot (1-M) \odot 4 $$
And inside $M'$ motion blur is applied. With the same mask multiplication, similar artifacts can be generated.

\paragraph{Training description:}
EfficientNet-b7 is used as the backbone network for all models, and no resizing and alignment is required. Model parameters are initialized by EfficientNet-b7 pre-trained on ImageNet. The batch size is set to 64. Adam is used for optimization with a learning rate of 0.00005, and after each epoch, the learning rate is decayed by 0.1. Weight decay is set to 0.0004, and models are trained for less than 3 epochs. For ternary classification,  $80\%$ of real images in the training set are  randomly selected and AIM is applied to those images on-the-fly during training.

In the training phase, the checkpoints of baseline models after 9500 iterations and 21250 iterations are chosen, multi-modal models after 10500 iterations and 20250 iterations are chosen, and multi-class model after 20000 iterations are chosen. After the first version of multi-class models is trained, it is noticed that a relative performance raise happen to the public testing set. Then an improved version of AIM is designed with more diversity of data augmentations at different levels, and multi-class model is fine-tuned using the checkpoint after 20000 iterations as pre-trained model, for 1500 iterations and 3750 iterations.

\paragraph{Testing description:}
Six EfficientNet-b7 models are trained under the aforementioned different training settings. Each model predicts independently, and the final results are generated using the mean value of all the predictions from the 6 models. 

\subsubsection{Generalization Analysis.}
To improve generalization, inspired by previous methods like Face X-Ray \cite{facexray} and SRM, a multi-modal, multi-class ensemble setting for multi-forgery detection is proposed. The innovations include: 1) a ternary classifier for both seen and unseen forgery attacks, 2) a cross-modality design for color learning and residual noise learning, and 3) a shadow-like artifact simulating method for face-swapping detection. In details,  two modality of input images are  utilized to extract features: RGB color modality and SRM filtered residual modality. The baseline model is trained in the provided training set and capable of detecting seen forgeries. Beyond that,  models using residual as input are trained  to enhance the textures of forgeries, and as a supervision for model to focus more on image noises rather than color information.

\subsection{Solution of the Third Place}
\begin{itemize}
    \item Solution title: \textit{More generalized DFD: Model matters, so does data representation}
    \item Team Name: \textit{TrustCV}
    \item Team members: \textit{Ying Guo, Yingying Ao, Pengfei Gao}
\end{itemize}
\subsubsection{General Method Description.}
\begin{figure}[h]
\centering
\includegraphics[width=\linewidth]{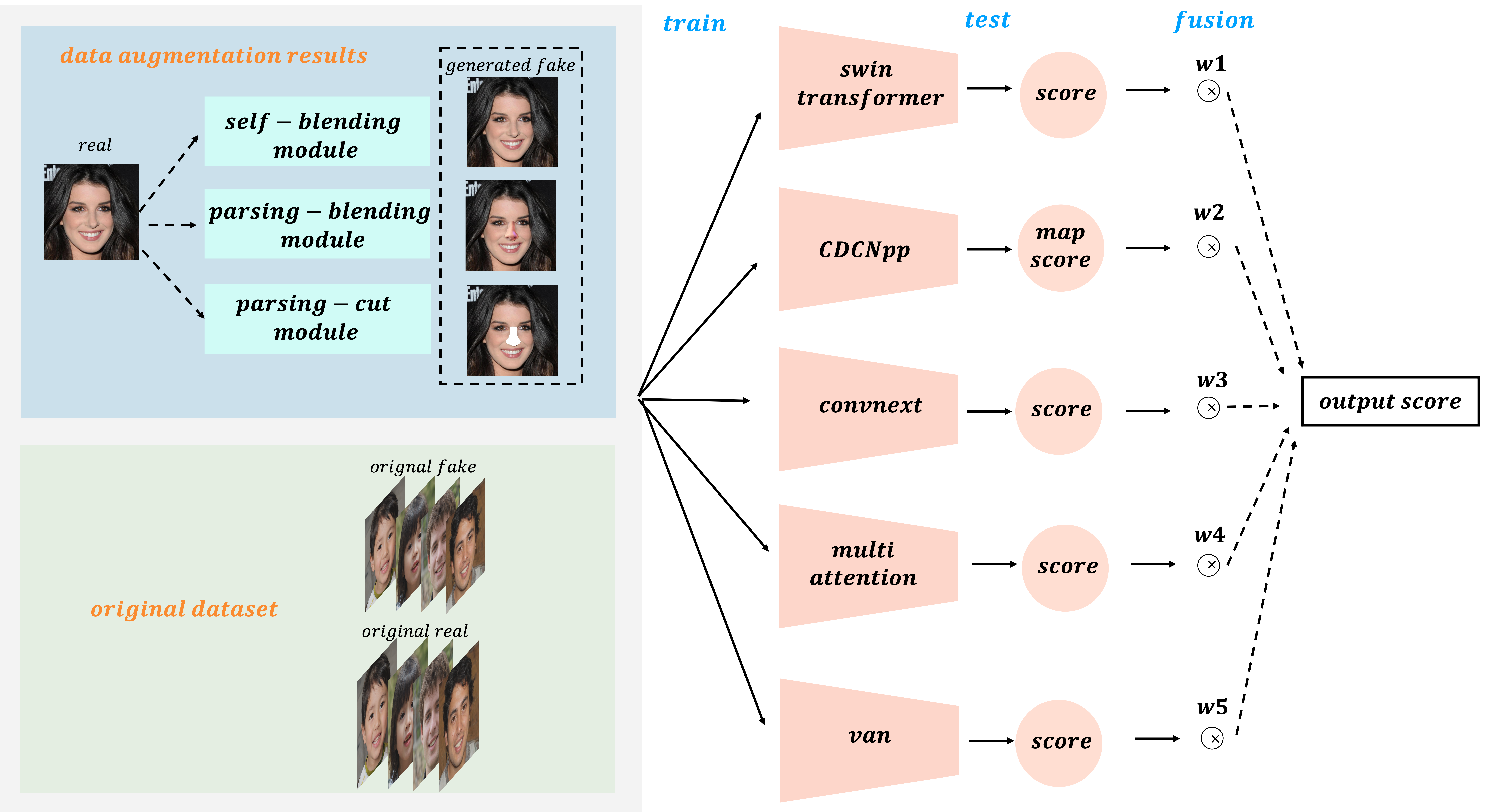}
\caption{\small The overall pipeline of general method.}
\label{fig:overall}
\end{figure}

To improve the generalization of forgery detection ability,  the solution mainly focuses on four aspects: (1) data augmentation, (2) model selection and data representation, (3) model fusion, and (4) detection ability testing. The main process of the method is as follows, and this pipeline is shown in Figure \ref{fig:overall}.

\textbf{Data augmentation}: First, two data augmentation methods are used to create more fake training data. The two data augmentation methods are called \emph{self-blending} and \emph{parsing blending/cut} respectively. The former aims to create a more common, general and robust forgery representation among different attack methods, so as to avoid over-fitting to a specific attack form, while the latter makes the model focus on more detailed and local changes of facial parts.

\textbf{Model selection \& Data representation}: Then, five models are  selected with different backbones, namely SwinTransformer \cite{swin}, CDCNpp \cite{cdcn}, ConvNext \cite{convnet}, Multi-Attention \cite{multi}, VAN \cite{van}, and  the models are trained on the training set composed of the original data and the data augmentation results. Among them, CDCNpp \cite{cdcn} is trained with the input of global and local patches respectively, and takes an image-grayscale based binary mask as the output feature representation; SwinTransformer \cite{swin} also uses two different data pre-processing methods. Thus, the diversity of data representation can be increased to improve the performance.

\textbf{Model Fusion}: After obtaining every single model,  these models are evaluated on the official validation set and their self-made fake dataset. According to the performance, a coarse-grained Particle Swarm Optimization (PSO) algorithm is used to search the weight of each model to integrate them. Then the weights are fine-tuned according to the bias of each model's ability to detect real or fake data. Finally the final predicted score of the image is obtained using the ensemble model.

\textbf{Detection ability testing}: To better test the generalization of the model, in addition to the official dataset,a fake dataset containing eight fake methods (i.e. FaceMorph, StarGANv2 \cite{stargan}, StyleGAN2 \cite{styleganv2}, FaceEdit \cite{faceedit}, HifiFace \cite{hififace}, SimSwap \cite{simswap}, FaceSwap2d, FaceSwp3d) are generated for model evaluation.

\subsubsection{Implementation details.}
\paragraph{Data augmentation:}
In the data pre-processing stage, two data augmentation methods are performed on the real images of the training set, which are called \emph{self-blending} and \emph{parsing blending/cut}, respectively.

\textbf{Self-blending:} Inspired by the work in \cite{selfblend}, they aim to create a more general and robust forgery representation to simulate common problems such as color mismatch, artifact and frequency inconsistency in different attack methods, so as to avoid over-fitting to a specific attack form and improve the generalization of the model. Specifically, given an initial real image $I$,  it is copied into two images, $I_{1}$ and $I_{2}$. The color and spatial transformations  are randomly selected for the two images (both transformations may be applied to the same or different images). For color transformation, the image channel, saturation, brightness, contrast, etc. are  changed, and downsampling and sharpening are conducted to change the color style and image frequency. For spatial transformation, on the one hand, the image is resized and shifted to get image $O_{2}$. On the other hand, based on the face parsing mask, the face region is  filled in  and the same affine transformation is  performed to obtain the face mask $M$ corresponding to $O_{2}$, and then  random \cite{facexray} and elastic \cite{elastic} deformation, Gaussian blur and transparency is performed to change  $M$ to obtain the final mask. Based on the mask,  the two transformed images are blended to obtain the output. In this way, through the random combination of different transformations, the generated fake images can cover many problems (color, edge, frequency and so on), and the self-blending also avoids the problem of landmark mismatch caused by the fusion of two different images.

\begin{figure}[h]
\centering
\includegraphics[width=\linewidth]{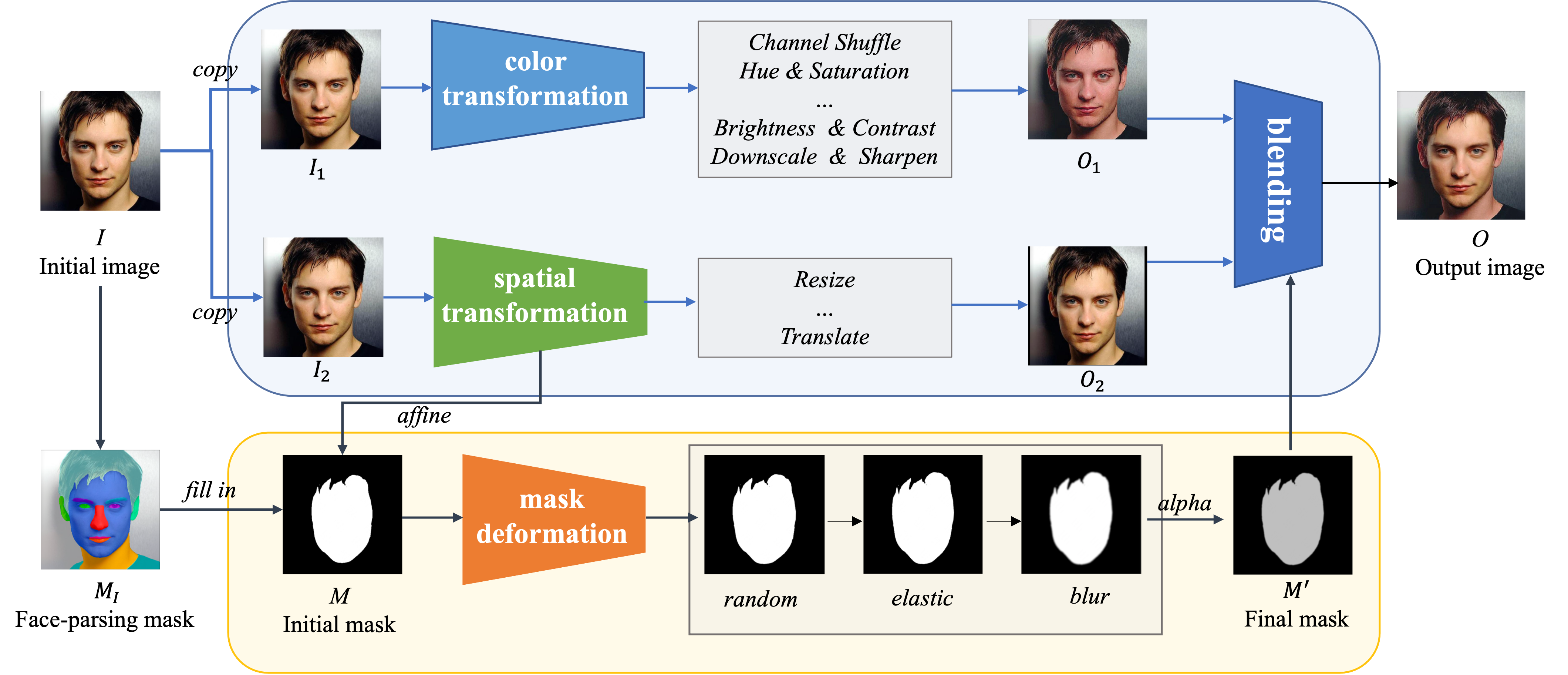}
\caption{\small The pipeline of self-blending in data augmentation.}
\label{fig:sf-blending}
\end{figure}

\begin{figure}[h]
\centering
\includegraphics[width=\linewidth]{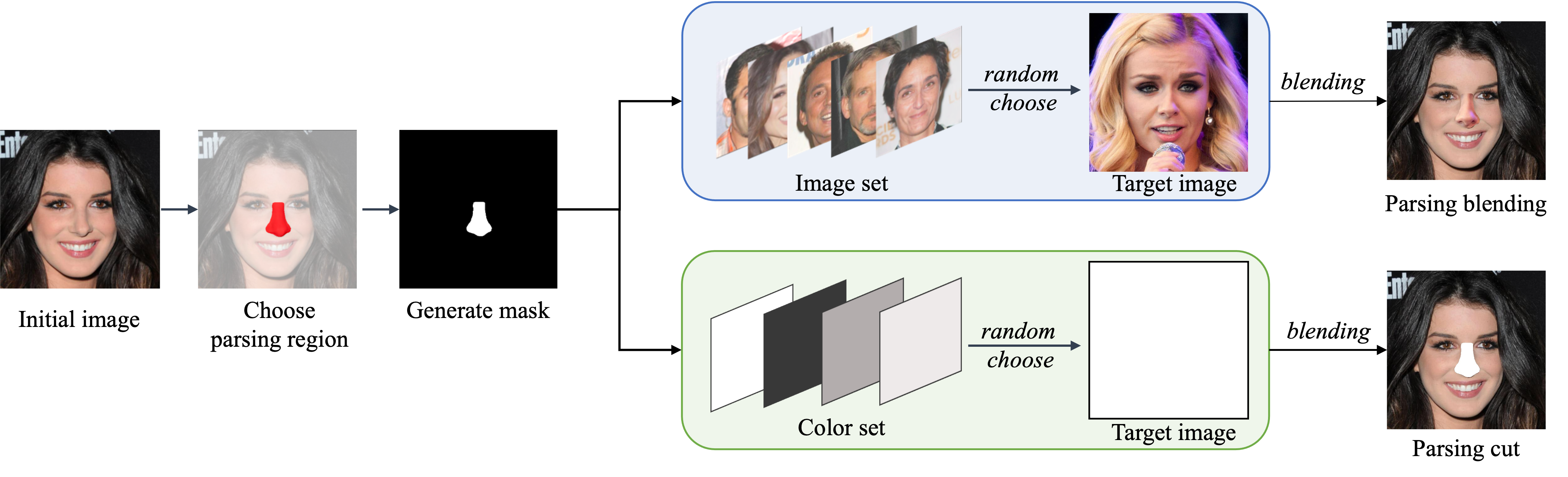}
\caption{\small The pipeline of parsing blending/cut in data augmentation.}
\label{fig:parsing}
\end{figure}

\textbf{Parsing blending/cut:} To make the model focus on more detailed and local changes of facial features and adapt to some fake samples in the Multi-FDC dataset,  blending and cutting augmentations are carried out  based on the face parsing results. The overall pipeline is shown in Figure \ref{fig:parsing}. First,  a local region based on the parsing results is  selected and the corresponding mask is generated. Then, on the one hand,  an image from the dataset is  randomly selected and mixed with the original image, which is called parsing blending; On the other hand, a solid color image of the same size is selected as the original image it is mixed so that the pixels in the selected area are changed to a fixed color, which is called parsing cut.

\paragraph{Training description:}
Totally 7 single models (5 types) are used for  further model ensemble. The training details of every single model are described as follows:

(1) SwinTransformer\_base \cite{swin}: The SwinTransformer is used as the base network. The input images are resized to $384\times 384$ and the model is initialized with the pre-trained ImageNet-1K model. Two versions of SwinTransformer are trained, which are different in data augmentation methods. The first one uses cut-out, coarse-dropout, etc., to force the network to focus on the forgery-specific discriminative information. The other one utilizes a variety of methods including image-compression, Gauss noise, shift-scale-rotate, etc., which improve the model's generalization.

(2) CDCNpp \cite{cdcn}: The Central Difference Convolutional Network \cite{cdcn} proposed in traditional face anti-spoofing task is used here. Two models are trained with different inputs.  The whole image is resized to $256\times 256$ as input for the first one. For the other, instead of training on the whole image,  random patch with size $224\times 224$ is used as input. All the models are supervised by gray-scale image mask.

(3) ConvNext \cite{convnet}: For ConvNext,  the base version is used, in which images are resized to $384\times 384$ as inputs. The pre-trained ImageNet-1k model is used.

(4) Multi-Attention \cite{multi} network: The Multi-Attention network proposed in CVPR 2021 is used, which consists of three import components: an Attention Module for generating multiple attention maps, a texture enhancement block, and a bilinear attention pooling for aggregating textural and semantic features. During  training the images are resized to $340\times 340$ as inputs, and the pre-trained EfficientNet-b4 \cite{tan2019efficientnet} on ImageNet-1K is used.

(5) VAN \cite{van}: For Visual Attention Network (VAN),  pre-trained VAN-Base model on ImageNet-1K is used. The images are resize to $384\times 384$ as the inputs to the model.

The image transformation strategies used by different models during the training process are summarize in Table \ref{tab:imgtrans}. They include Image Compression, Cut Out, Gauss Noise, Horizontal Flip, Random Brightness Contrast, ToGray, Shift/Scale/Rotate (S/S/R), and ColorJitter.

\begin{table}[h]   
\begin{center}   
\label{table:1} 
\resizebox{\linewidth}{!}{%
\begin{tabular}{c|c|c|c|c|c}   
\toprule[1.2pt]   \textbf{Methods} & \textbf{Swin} & \textbf{Convnext} & \textbf{CDCNpp} & \textbf{Multi-Att} & \textbf{VAN}\\   
\hline   Img Compression          & \checkmark    & \checkmark        &                 & \checkmark           &  \\
\hline   Cut Out                  & \checkmark    & \checkmark        & \checkmark      & \checkmark        &  \checkmark\\  
\hline   Gauss Noise             & \checkmark    &  \checkmark       &                 &   \checkmark       &  \\
\hline   Horizontal Flip          & \checkmark    & \checkmark        & \checkmark        & \checkmark     & \checkmark \\ 
\hline   Bright\&Contrast        & \checkmark     & \checkmark       &                  &  \checkmark       &  \\ 
\hline   ToGray                 & \checkmark     & \checkmark        &                  &  \checkmark        &  \checkmark\\ 
\hline   S/S/R                    & \checkmark     & \checkmark        &                  &  \checkmark       &  \\ 
\hline   ColorJitter             &  \checkmark    &                   & \checkmark      &                     & \checkmark \\
\bottomrule[1.2pt]   
\end{tabular}
}
\end{center}
\caption{Image transforms in the training stage of the adopted models}\label{tab:imgtrans}
\end{table}

\paragraph{Testing description:}
In the testing stage, to improve the accuracy and robustness of single models, the final score of an image is the average of results with and without image flip process. The output scores of different methods will be assigned with different weights, searched by the Particle Swarm Optimization (PSO) algorithm, which further enhances the performances of the proposed method.

\subsubsection{Generalization Analysis.}
To ensure the generalization ability of the method,  efforts in the following three aspects are made: (1) self-blending and parsing blending/cut augmentation methods are used. The former focuses on the common problems of different attack methods (~\emph{e.g.} color mismatch, artifact and frequency inconsistency), and forces the model to learn more general and robust feature representations, and the latter makes the model to focus on more detailed and local changes of facial features. (2) By training different models, feature representations of different forms are obtained, which include RGB(ConvNext, SwinTransformer, VAN), mask (CDCNpp) and texture maps (Multi-Attention), and different scales including global (SwinTransformer, Multi-Attention, VAN, ConvNext) and local(CDCNpp) aspects. As a result, the model can obtain more diverse information to improve its generalization. (3) For better generalization evaluation,  a self-made forgery dataset using eight forgery methods different from the official dataset is constructed, including FaceMorph, StarGANv2 \cite{stargan}, StyleGAN2 \cite{styleganv2}, FaceEdit \cite{faceedit}, HifiFace \cite{hififace}, SimSwap \cite{simswap}, FaceSwap2d, FaceSwp3d.

\section{Discussions}
From the technical reports, we can see that most solutions use data augmentation, data extension and image blending. We can also see that participants are making attempts to model unseen forgery types, perform feature representation, data representation, and model ensemble~\emph{etc.}

\begin{figure}
  \centering
    \includegraphics[width=0.3\textwidth]{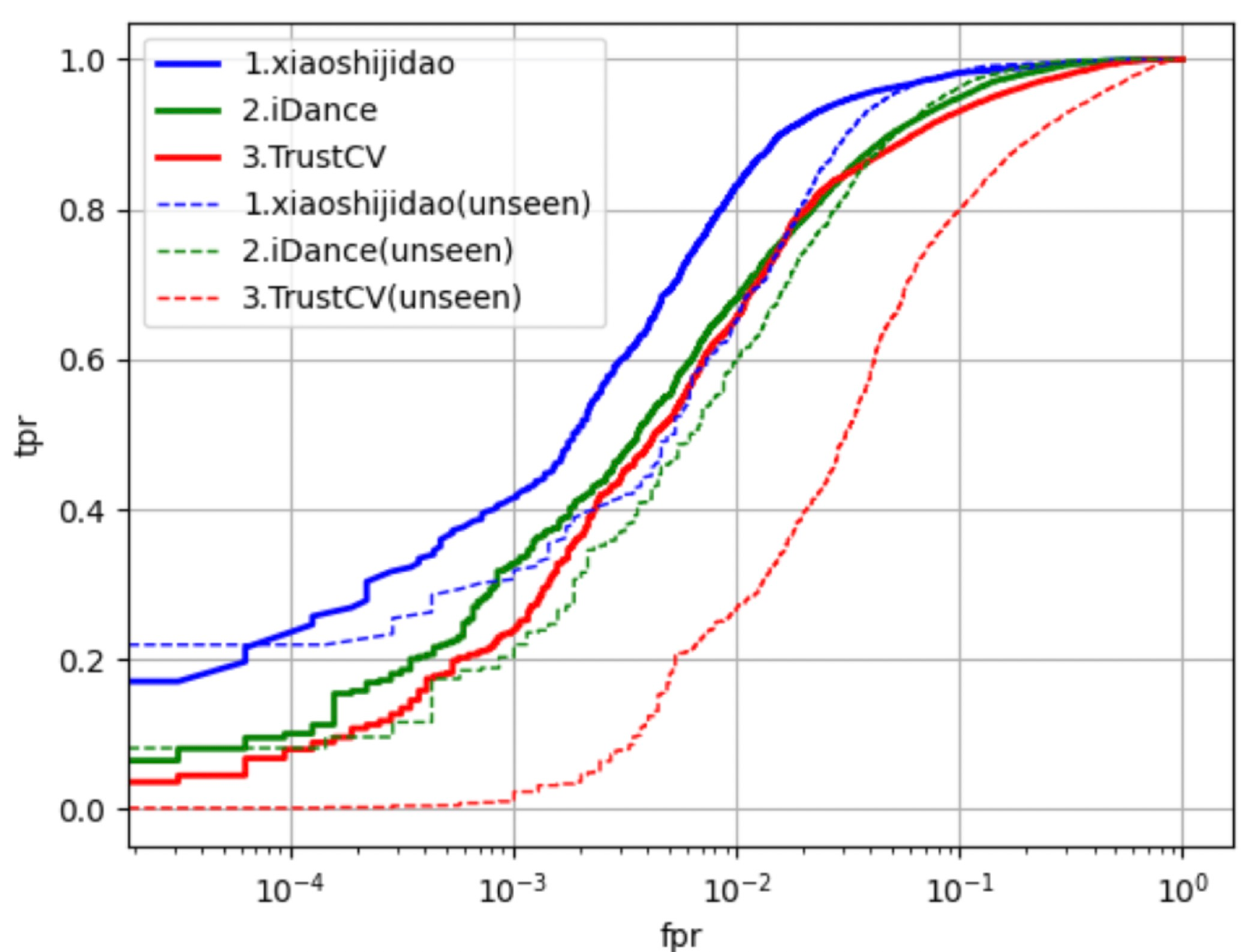}
    \caption{ROC curve of Top 3 teams in terms of seen and unseen types of forgery on the test set. }
  \label{fig:unseen}
\end{figure}

Although the participants achieve relatively high performance in terms of AUC, it doesn't mean the image forgery detection problem is a solved problem. To demonstrate this, we perform a deep analysis of the submitted solutions. The results of the analysis are shown in Figure~\ref{fig:unseen}. On the one hand, the performance gap on seen forgery types and unseen forgery types is sometimes very significant, especially when FPR is low. The performance gap can be up to $50\%$.  On the other hand, the TPR on low FPR is not satisfactory. Here FPR measures the rate of normal images classified as forged images,~\emph{i.e.} the rate of normal users being disturbed by the algorithm. In real-world applications, the base number of  users is usually big, which requires that the normal user disturbance rate, ~\emph{a.k.a} FPR, should be very low,~\emph{e.g.} $1/1000$ or even $1/10000$. At such low FPR, the TPR, which measures the rate of forged images classified as forged images, is not usable yet. For example, TPR at FPR=$1/1000$ is only about $40\%$, even for the top ranking team. This means that more than half of the forgery images are not detected.

From the analysis, we can see that more research work is needed to close the performance gap of seen and unseen types. Moreover, how to improve TPR at low FPR is also a problem that needs more research attentions. 

\section{Conclusion}
\vspace{-3mm}
In this Multi-Forgery Detection Challenge we formulate the problem of multi-forgery detection. During the challenge we have seen  some successful attempts to the problem,  such as unreasonable effectiveness of data, simulation of unseen types of forgery, and variety of models as inductive biases~\emph{etc}. From the other side we notice that explicit modeling of unseen forgery types and architecture design for forgery detection task is more or less missing. From the challenge, we can also see some promising future directions in the field of image forgery detection.  Large-scale and diverse datasets, generalizability to unseen forgery types, and fast adaptation to certain forgery types~\emph{etc.} are promising directions that can be further explored.

\newpage

{\small
\bibliographystyle{splncs04}
\bibliography{references}
}

\end{document}